\documentclass[conference]{IEEEtran}

\usepackage[T1]{fontenc}
\usepackage{newtxtext}

\IEEEoverridecommandlockouts

\usepackage{cite}
\usepackage{amsmath,amssymb,amsfonts}
\usepackage{algorithmic}
\usepackage{graphicx}
\usepackage{textcomp}
\usepackage{xcolor}
\usepackage{booktabs}
\usepackage{tabularx}

\usepackage{xurl}
\usepackage[hidelinks]{hyperref}
\usepackage{mdframed}
\usepackage[most]{tcolorbox}     

\def\BibTeX{{\rm B\kern-.05em{\sc i\kern-.025em b}\kern-.08em
    T\kern-.1667em\lower.7ex\hbox{E}\kern-.125emX}}

\begin{document}

\title{
A Hybrid Deterministic Framework for Named Entity Extraction in Broadcast News Video
\thanks{
This work is part of the project \emph{Transparent Media Monitoring and Understanding (TMMU)}.
It is financed by Xjenza Malta and the Malta Digital Innovation Authority (MDIA) for and on behalf of the Foundation for Science and Technology through the FUSION: R\&I Thematic Programmes, Digital Technologies Programme.
}
}

\author{
\IEEEauthorblockN{Andrea Filiberto Lucas}
\IEEEauthorblockA{
Dept. of Artificial Intelligence\\
University of Malta\\
Msida, Malta\\
andrea.f.lucas@um.edu.mt
}
\and
\IEEEauthorblockN{Dylan Seychell}
\IEEEauthorblockA{
Dept. of Artificial Intelligence\\
University of Malta\\
Msida, Malta\\
dylan.seychell@um.edu.mt
}
}
\maketitle

\begin{abstract}
The growing volume of video-based news content has heightened the need for transparent and reliable methods to extract on-screen information. Yet the variability of graphical layouts, typographic conventions, and platform-specific design patterns renders manual indexing impractical. This work presents a comprehensive framework for automatically detecting and extracting personal names from broadcast and social-media-native news videos. It introduces a curated and balanced corpus of annotated frames capturing the diversity of contemporary news graphics and proposes an interpretable, modular extraction pipeline designed to operate under deterministic and auditable conditions.

The pipeline is evaluated against a contrasting class of generative multimodal methods, revealing a clear trade-off between deterministic auditability and stochastic inference. The underlying detector achieves 95.8\% mAP@0.5, demonstrating operationally robust performance for graphical element localisation. While generative systems achieve marginally higher raw accuracy (F1: 84.18\% vs 77.08\%), they lack the transparent data lineage required for journalistic and analytical contexts. The proposed pipeline delivers balanced precision (79.9\%) and recall (74.4\%), avoids hallucination, and provides full traceability across each processing stage. Complementary user findings indicate that 59\% of respondents report difficulty reading on-screen names in fast-paced broadcasts, underscoring the practical relevance of the task. The results establish a methodologically rigorous and interpretable baseline for hybrid multimodal information extraction in modern news media.
\end{abstract}


\begin{mdframed}[
    linewidth=0.4pt,
    roundcorner=4pt,
    innertopmargin=6pt,
    innerbottommargin=6pt,
    innerleftmargin=8pt,
    innerrightmargin=8pt
]
\centering
\small
\textit{This paper has been selected for publication in the\\
2026 IEEE Conference on Artificial Intelligence (CAI).}
\end{mdframed}

\section{Introduction}
The contemporary media ecosystem produces vast volumes of video content that increasingly exceed the capacity of both audiences and analysts to process effectively \cite{1,2}. With short-form news formats becoming dominant across broadcast and online platforms \cite{3}, the need for systems capable of automatically extracting salient contextual information has intensified \cite{Kenely}. Personal names remain central identifiers within news narratives, yet they are typically embedded within dynamic graphical elements whose typography, spatial layout, and temporal variability hinder accurate machine interpretation \cite{8,9,10,11,12}. A survey conducted as part of this study found that 59\% of respondents experience difficulty reading on-screen names, underscoring an accessibility gap that motivates automated, reliable extraction mechanisms.

This study investigates the automatic detection and recognition of personal names presented within broadcast video graphics, extending prior work in visual information extraction \cite{8,9,10,11,12}. Building on earlier systems \cite{15}, it introduces a generalisable approach to visual named-entity identification that integrates computer vision (CV), optical character recognition (OCR), and linguistic entity modelling in a deterministic and auditable pipeline \cite{16}. In parallel, the study examines the performance characteristics of contemporary generative multimodal approaches, highlighting the methodological trade-off between transparent, rule-governed inference and stochastic generative reasoning.

The main contributions of this work are as follows:
\begin{enumerate}
    \item The creation of the \textbf{News Graphics Dataset (NGD)}, a curated corpus of annotated frames capturing diverse graphical conventions across broadcast and social-media-native news sources.
    \item The development of an interpretable \textbf{visual name extraction pipeline} designed around deterministic processing stages, enabling transparent data lineage and robust identification of personal names in heterogeneous graphical environments.
    \item A systematic comparative analysis \textbf{between deterministic and generative multimodal approaches}, quantifying the empirical trade-offs between auditability, reliability, and raw extraction accuracy.
\end{enumerate}

\section{Related Work}
Automated extraction of textual information from news video graphics builds on prior work in object detection (OD), OCR, named entity recognition (NER) and more recent multimodal learning. Elements such as lower-thirds, tickers, and headlines act as visual anchors that convey names, roles, and locations \cite{13,17,18}. Their variability in typography, colour, layout, and motion across broadcasters presents significant challenges for detection and interpretation \cite{20,21}. Overlapping elements, semi-transparent layers, and differences in stylistic design complicate segmentation and recognition, which highlights the need for domain-specific datasets and robust pipelines that can accommodate the temporal and visual diversity of broadcast media.

Existing corpora for video analysis provide broad coverage of broadcast material but generally lack the fine-grained annotations required to study graphical overlays at the level of detail demanded by visual text extraction tasks. Resources such as the TV News Archive \cite{42} and the datasets introduced by Lee et al.\ \cite{43} and Guha et al. \cite{dataset-2} incorporate large volumes of news footage but omit explicit bounding-box annotations and do not capture the stylistic diversity characteristic of contemporary on-screen graphics. The \textbf{News Graphics Dataset (NGD)} was developed to address this gap by providing a curated set of precisely annotated frames that reflect the typographic, spatial, and thematic variability found across Maltese, international, and social-media-native news sources. This enables systematic, reproducible evaluation of methods for extracting and interpreting textual information embedded within dynamic graphical elements.

In visual element detection, single-stage convolutional frameworks such as YOLO have proved effective due to their unified localisation and classification design \cite{22,23,25}. Later versions, including YOLOv12, incorporate attention mechanisms and multi-scale prediction, improving accuracy in dense broadcast scenes \cite{b27,b54}. Two-stage detectors such as Faster R-CNN \cite{46,49} provide strong precision but require higher computational cost, which limits suitability for real-time or resource-constrained deployments. Efficiency is further improved through frame-sampling and deduplication techniques based on perceptual hashing or feature descriptors \cite{29,30,47}.

OCR forms a central component in video-text extraction. Preprocessing techniques such as greyscale conversion, adaptive thresholding, and CLAHE improve legibility, although stylised, low-resolution, or animated text remains difficult to recognise reliably \cite{31,59}. Transparent open-source engines, including Tesseract \cite{34,60}, offer adaptability for domain-specific tuning, while high-performing cloud-based systems have been shown to provide improved accuracy at increased computational and privacy cost \cite{68}. Downstream, NER models identify and validate person names within noisy OCR outputs. Transformer-based architectures, including BERT and GLiNER, consistently outperform earlier CRF or LSTM approaches and support multilingual generalisation \cite{37}, although OCR noise continues to be a major limiting factor \cite{71}.

Recent multimodal generative systems extend these capabilities by jointly analysing visual and textual cues \cite{83,89}. These models demonstrate flexibility in interpreting complex graphical layouts, yet their stochastic and non-deterministic behaviour raises concerns regarding factual consistency, reproducibility, and transparency \cite{40}. As a result, hybrid approaches that combine deterministic OD, OCR, and NER pipelines with multimodal reasoning or validation have emerged as a practical compromise. These methods integrate the interpretability of deterministic components with the contextual adaptability of generative models in order to improve accuracy and robustness across heterogeneous broadcast environments.

\section{Methodology}
This study employs a multi-stage methodological framework encompassing dataset construction, model training, and comparative evaluation. The NGD was developed to capture the stylistic and structural variability of broadcast and social-media news overlays. A set of detectors based on the YOLO family of models was trained on this dataset and serves as the foundation for the Accurate Name Extraction Pipeline (ANEP). The pipeline is evaluated against contemporary generative multimodal systems to establish a fair benchmark for robustness and reliability, using two distinct multimodal baselines. ANEP specialises the broader named-entity extraction task by focusing explicitly on personal names that appear within news-graphic content.

The NGD is publicly available via Roboflow\footnote{\url{https://universe.roboflow.com/ict3909-fyp/news-graphic-dataset}}, while the ANEP implementation is accessible on GitHub\footnote{\url{https://github.com/AFLucas-UOM/Accurate-Name-Extraction}}.

\subsection{The News Graphics Dataset (NGD)}
The NGD comprises 300 videos sourced from local, international, and social-media news outlets, with each provider capped at 12.5\% for balanced representation. Frames were selected using temporal segmentation and perceptual similarity analysis, followed by manual verification to ensure diversity and annotation quality. The final dataset contains 1,500 frames and 4,749 region-level annotations across six categories: \emph{Breaking News Graphics}, \emph{Digital On-Screen Graphics}, \emph{Lower Thirds}, \emph{Headlines}, \emph{Tickers}, and \emph{Other News Graphics}. All images were standardised to 640$\times$640~px and augmented with controlled adjustments to brightness, exposure, and noise. A stratified 93\% / 4\% / 3\% train-validation-test split was applied after frame extraction. Although partitioning occurred at the frame level, temporal segmentation and perceptual deduplication substantially reduce near-duplicate frames, and broadcast graphics typically vary enough to minimise the risk of train-test leakage.

\subsection{YOLO-based Object Detection}
Variants from the YOLO family of detectors, including v8, v11, v12, and NAS, were evaluated on the NGD. YOLOv12 produced the highest mean Average Precision and was therefore selected for downstream processing. Models were trained from scratch at 640$\times$640~px using cosine-annealed learning rates and early stopping. The attention-oriented design of YOLOv12 improved the localisation of small and overlapping graphical elements, which commonly appear in broadcast video. The resulting detections were used to define regions of interest for text extraction.

\subsection{The Accurate Name Extraction Pipeline (ANEP)}
ANEP is organised into five sequential stages:
\begin{enumerate}
    \item \textbf{Frame Processing}, which samples frames at 1~FPS and removes redundancies through hashing;
    \item \textbf{Graphic Detection and Pre-processing}, which applies YOLOv12 followed by contrast enhancement and adaptive thresholding;
    \item \textbf{Text Extraction}, which performs OCR and filters outputs according to confidence;
    \item \textbf{Named-Entity Recognition and Validation}, which combines a fine-tuned transformer-based NER model with a supplementary linguistic pipeline supported by heuristic filters; and
    \item \textbf{Name Clustering and Timeline Generation}, which merges variants using fuzzy, Jaccard, and embedding-based similarity to produce canonical names and temporal occurrence profiles.
\end{enumerate}

\begin{figure}[h]
\centering
\includegraphics[width=0.48\textwidth]{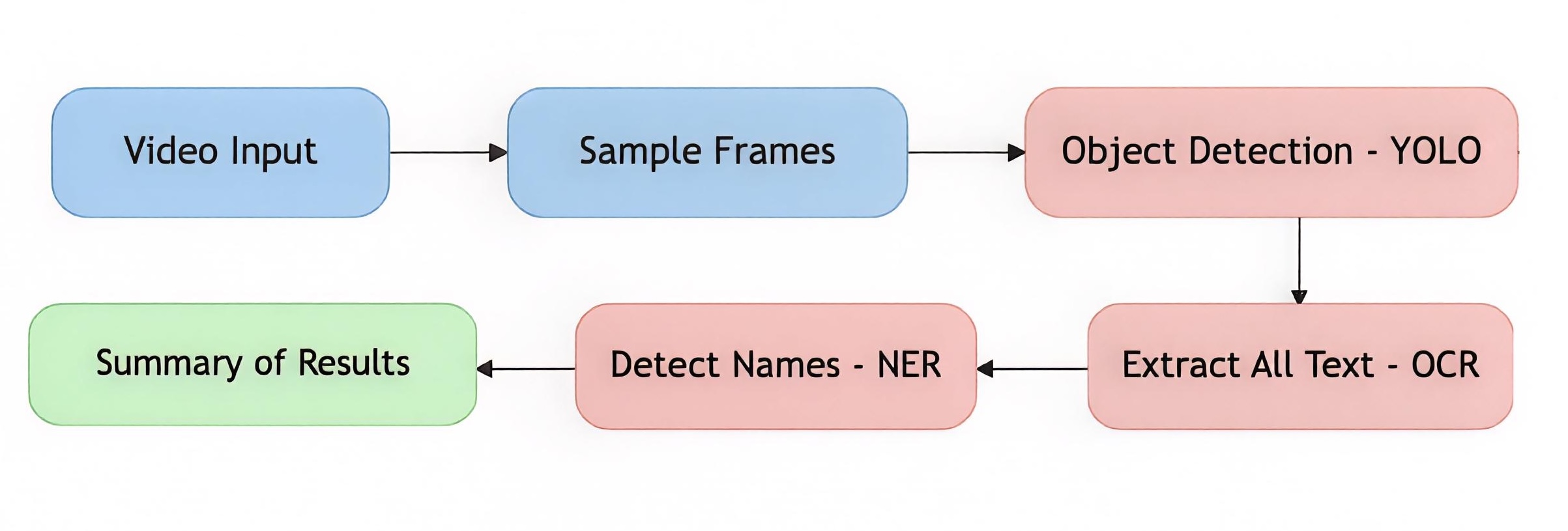}
\caption{Simplified workflow of ANEP, illustrating the sequential progression from video ingestion to object detection, text extraction, name recognition, and final result synthesis.}
\label{fig:anep_flow}
\end{figure}

This design provides access to transparent intermediate outputs and supports interpretability at each processing stage, which facilitates diagnostic evaluation and domain-specific optimisation.

\subsection{The GenAI Multimodal Pipelines}
Two generative multimodal pipelines were implemented for comparative evaluation. The first employed Gemini~1.5~Pro, and the second used LLaMA~4~Maverick. Both systems performed multimodal extraction using structured prompts that requested only real-world personal names. Video frames were deduplicated through perceptual hashing and submitted as base64-encoded images for text detection and reasoning. Outputs were returned in JSON format and evaluated with the same metrics applied to ANEP.

The Gemini pipeline provided faster inference and strong precision, while LLaMA~4~Maverick showed comparable reasoning capabilities. However, both systems exhibited variable latency and limited reproducibility. In contrast, ANEP offers explicit control, consistent outputs, and verifiable saliency at each stage of its OD, OCR, and NER pipeline.

\section{Evaluation of Results}
The evaluation employed standard metrics for classification, object detection, and name extraction in order to assess the accuracy, robustness, and efficiency of the YOLO-based models, ANEP, and the generative multimodal baselines. Quantitative analysis and comparative benchmarking provide a critical assessment of each method, identifying strengths as well as limitations.

\begin{table*}[t]
\centering
\footnotesize
\renewcommand{\arraystretch}{1.1}
\setlength{\tabcolsep}{3pt}
\caption{Performance comparison of YOLO-based object detection models trained on the NGD.  \\
The best-performing configuration, \textbf{YOLOv12\textit{(m)}}, achieved the highest overall accuracy across all metrics.}
\label{tab:yolo-comparison}
\begin{tabular*}{\textwidth}{@{\extracolsep{\fill}}lccccc}
\toprule
\textbf{Model} & \textbf{Precision (\%)} & \textbf{Recall (\%)} & \textbf{mAP@0.5 (\%)} & \textbf{mAP@0.5:0.95 (\%)} & \textbf{Epochs} \\
\midrule
\multicolumn{6}{c}{\textbf{Locally Trained Models}} \\
\midrule
\textbf{YOLOv12\textit{(m)}} & \textbf{93.9} & \textbf{93.5} & \textbf{95.8} & \textbf{88.7} & \textbf{102} \\
YOLOv8\textit{(m)} & 92.6 & 86.9 & 93.7 & 75.2 & 47 \\
\midrule
\multicolumn{6}{c}{\textbf{Externally Trained Configurations}} \\
\midrule
YOLOv12\textit{(n)} & 91.6 & 90.8 & 93.8 & 85.4 & 120 \\
YOLOv11\textit{(n)} & 91.2 & 90.4 & 93.1 & 84.9 & 100 \\
YOLOv12\textit{(n)} (Reflect) & 91.4 & 85.7 & 91.8 & 80.4 & 72 \\
YOLO\text{-}NAS\textit{(n)} & 85.1 & 84.3 & 91.0 & 61.0 & 51 \\
\bottomrule
\end{tabular*}
\end{table*}

\subsection{Comparative Performance Analysis of OD Models}
Table~\ref{tab:yolo-comparison} summarises the performance obtained across representative training runs. Externally trained configurations based on YOLOv12(n) achieved strong results, including 93.8\% mAP@0.5, 91.6\% precision, and 90.8\% recall. The highest overall performance, however, was obtained by the locally trained YOLOv12(m), which reached 95.8\% mAP@0.5 with 93.9\% precision and 93.5\% recall. The locally trained YOLOv8(m) delivered a comparable mAP@0.5 of 93.7\% but showed reduced recall at 86.9\%, indicating lower consistency in detecting all target instances. YOLOv11(n) achieved competitive recall of 90.4\% but slightly lower localisation accuracy at 93.1\% mAP@0.5. The YOLOv12(n) Reflect variant underperformed at 91.8\% mAP@0.5, and YOLO-NAS(n) provided the weakest generalisability, with 85.1\% precision and 84.3\% recall.

\begin{figure}[h]
    \centering
    \includegraphics[width=0.5\textwidth]{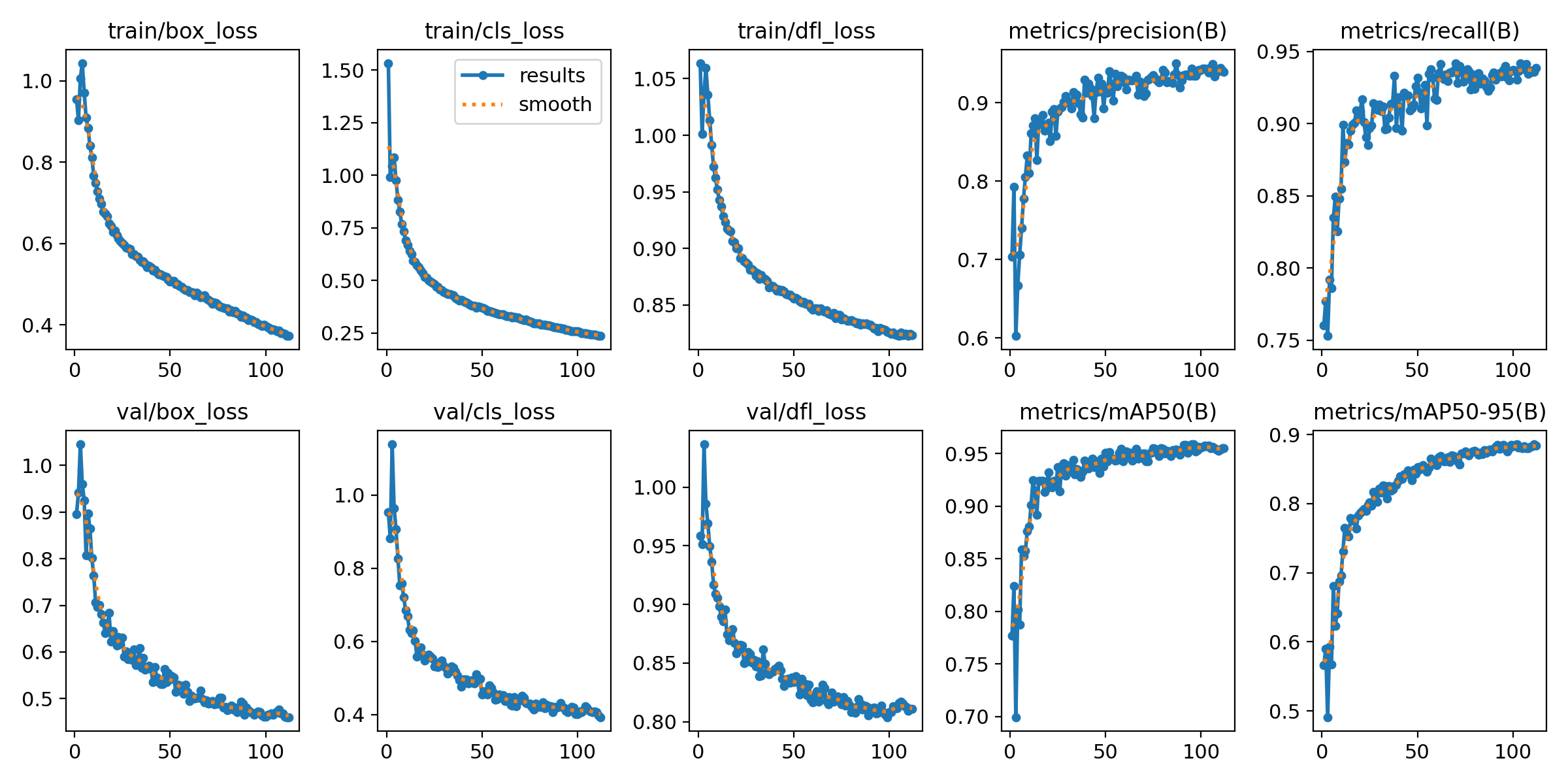}
    \caption{Training and validation curves for YOLOv12(m), showing box, classification, and object loss convergence alongside precision, recall, mAP@0.5, and mAP@0.5:0.95 metrics. The results demonstrate stable optimisation and consistent performance improvements across epochs, indicating robust generalisation and well-balanced precision-recall behaviour.}
    \label{fig:yolo-curves}
\end{figure}

The superior performance of YOLOv12(m) highlights the benefits of increased model capacity and training that is tailored to the characteristics of the NGD. Its 95.8\% mAP@0.5 demonstrates operationally robust performance for detecting news graphics and supports the model’s suitability for practical deployment. The corresponding loss and confidence curves in Figure~\ref{fig:yolo-curves} show stable convergence and a consistent balance between precision and recall throughout training.

\subsection{Visual Analysis of Model Attention and Generalisation}
\begin{figure}[h]
    \centering
    \includegraphics[width=0.5\textwidth]{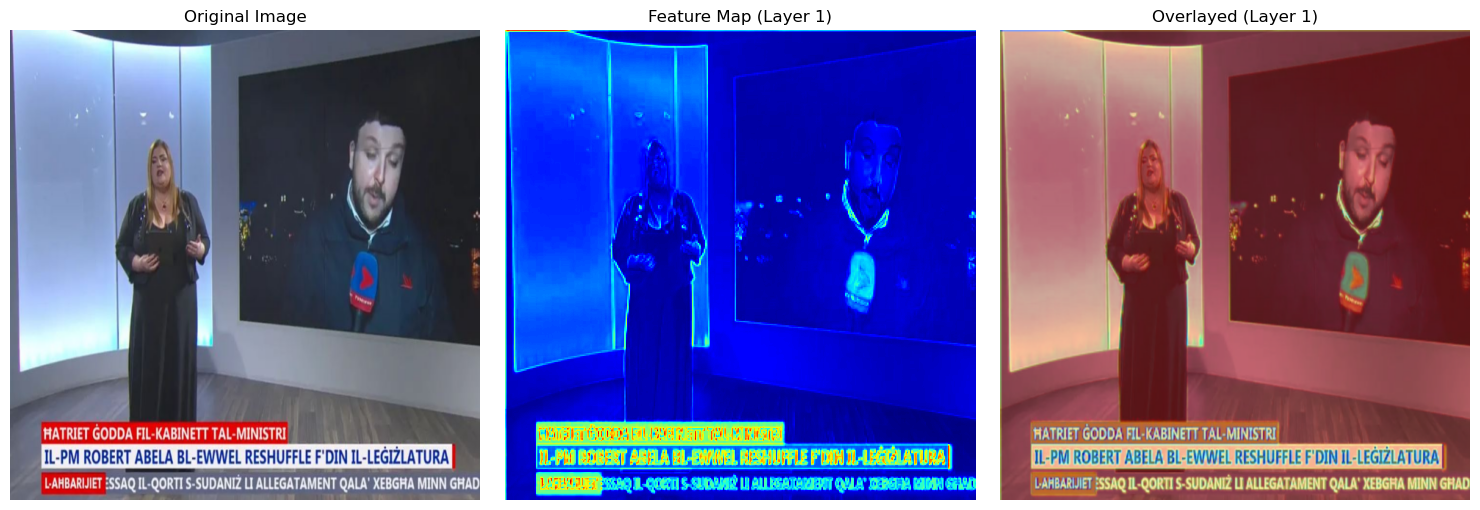}
    \caption{Grad-CAM visualisation for a representative frame illustrating the original input (left), activation map (centre), and overlay (right) derived from Layer~1 of YOLOv12(m), highlighting strong focus on text-rich regions.}
    \label{fig:gradcam}
\end{figure}

Visual interpretation methods were used to examine the internal behaviour and generalisation capability of the detector. Grad-CAM analysis (Figure~\ref{fig:gradcam}) showed strong activation in text-dense regions, particularly within lower-third graphics and tickers, indicating that the model captured spatial and semantic features relevant to the task.

\begin{figure}[h]
    \centering
    \includegraphics[width=0.5\textwidth]{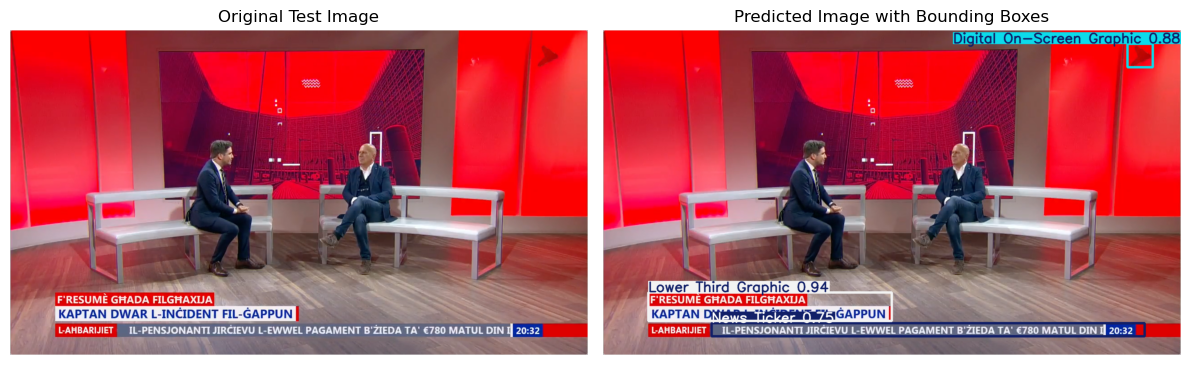}
    \caption{Sample detection from a TVM news broadcast demonstrating precise localisation of all graphical overlays and accurate bounding across all graphics.}
    \label{fig:tvm-sample}
\end{figure}

\begin{figure}[h]
    \centering
    \includegraphics[width=0.5\textwidth]{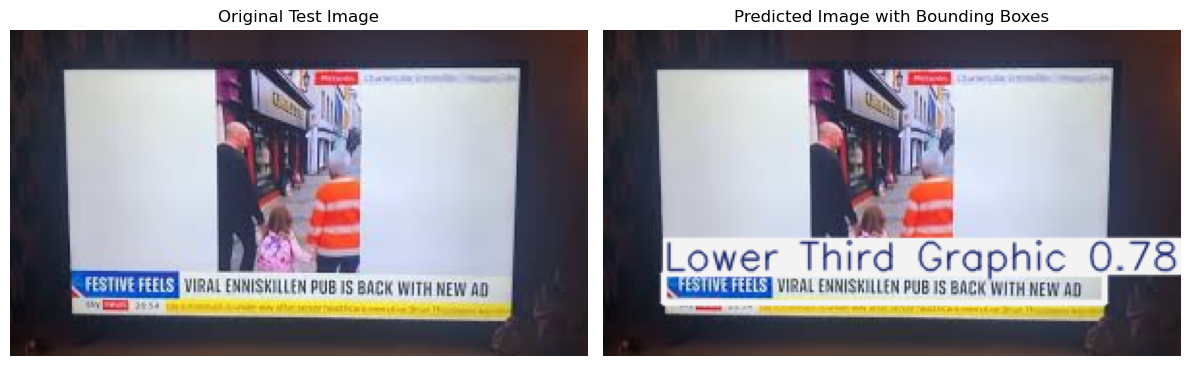}
    \caption{Detection example from a Sky News frame photographed from a television display. Despite glare and compression artefacts, the model successfully identified the principal graphical regions.}
    \label{fig:skynews-sample}
\end{figure}

Generalisation was assessed using previously unseen broadcast material. As illustrated in Figure~\ref{fig:tvm-sample}, the detector successfully identified graphical overlays across multiple formats. It also maintained reliable performance under more challenging conditions, such as photographed screens with glare (Figure~\ref{fig:skynews-sample}), although occasional merging occurred between adjacent elements including tickers and lower-thirds.

These results indicate that the detector can transfer effectively to new broadcast domains without additional training, which is essential for large-scale and continuously evolving media monitoring environments.

\subsection{Comparative Name Extraction Performance}
Evaluation was conducted against a manually constructed ground truth listing all named individuals in the test video, with frame-level timestamps for first and last appearances. System outputs were aligned using case-insensitive matching with fuzzy tolerance to mitigate OCR variation, and correctness was determined by temporal overlap. Precision, recall, F1 score, and runtime were computed to assess accuracy and efficiency. Runtimes were measured on a MacBook Air M2, with all methods evaluated on the same machine at a uniform 1~FPS sampling rate to ensure comparability. As shown in Table~\ref{tab:name-extraction-performance}, the results provide a comprehensive comparison across the three pipelines.

\begin{table}[t]
    \centering
    \footnotesize
    \renewcommand{\arraystretch}{1.1}
    \setlength{\tabcolsep}{5.5pt}
    \caption{Comparative performance of name extraction models.}
    \label{tab:name-extraction-performance}
    \begin{tabular}{lcccc}
    \toprule
    \textbf{Model} & \textbf{Precision} & \textbf{Recall} & \textbf{F1 (\%)} & \textbf{Runtime (s)} \\
    \midrule
    ANEP & 79.92 & 74.44 & 77.08 & 542.15 \\
    Gemini~1.5~Pro & \textbf{93.33} & \textbf{76.67} & \textbf{84.18} & \textbf{94.68} \\
    LLaMA~4~Maverick & 66.67 & 50.00 & 57.14 & 140.18 \\
    \bottomrule
    \end{tabular}
\end{table}

\subsubsection{Quantitative Results and Discussion}
The generative multimodal baseline built on the Gemini~1.5 Pro model achieved the strongest overall performance, with 93.33\% precision, 76.67\% recall, and an F1 score of 84.18\%. Its average runtime of 94.68~seconds indicates favourable computational efficiency and suggests suitability for near real-time monitoring. The model’s multimodal reasoning contributed to reliable extraction across varied graphical layouts. Although accuracy was high, the approach functions as a black-box system and does not expose intermediate representations, which limits interpretability and traceability.

ANEP achieved balanced precision (79.92\%) and recall (74.44\%), achieving an F1 score of 77.08\%. Performance was primarily limited by temporal misalignment and OCR-induced string variation. Its average runtime of 542.15 seconds arises from the sequential multi-stage pipeline, including 1 FPS frame sampling, deduplication, OCR, transformer-based NER, and clustering. The clustering and alias-resolution module is a major contributor due to its pairwise similarity computations. Despite this overhead, the modular design provides transparent intermediate outputs that support diagnostic analysis and auditability in journalistic and forensic contexts.

The LLaMA-based baseline produced the weakest results, with recall limited to 50.00\% and an F1 score of 57.14\%. Its average inference time of 140.18~seconds was lower than that of ANEP, but extraction was hindered by frequent misidentifications and omissions. The model struggled with overlapping or visually degraded text, indicating that general-purpose language models are not sufficiently specialised for structured visual extraction tasks without targeted adaptation.

Overall, the results highlight a trade-off between accuracy, transparency, and computational cost. These findings reinforce the continued relevance of modular and transparent pipelines for applications that require verifiable and accountable information extraction.

\subsubsection{Error Analysis and Limitations}
Analysis of the system outputs identified several recurring sources of error. Within ANEP, inconsistencies in named-entity recognition were common for surnames that appear independently as English lexical items, such as “Swift” or “Trump,” and for names that include diacritics, such as Ċensu, Żahra, or Ġużeppi. These issues reflect the sensitivity of OCR to orthographic variation and the limited representation of certain regional names within widely used training corpora. The alias resolution module successfully merged straightforward variants, for example “Donald Trump” and “President Trump,” but it frequently failed to reconcile indirect references or heavily distorted OCR outputs, such as “Mickey Rourke” versus “Brother Rourke.”

The LLaMA-based baseline was constrained by the context window of the underlying model, which reduced performance on longer video segments. Once inputs exceeded approximately five minutes, truncation effects caused weaker temporal consistency and hindered the tracking of named individuals, limiting suitability for continuous broadcast monitoring.

All evaluated pipelines exhibited higher error rates in scenes containing overlapping graphics, rapid shot changes, or stylised visual layouts. OCR artefacts often propagated to downstream components and produced both missed detections and false positives during entity matching. The absence of additional modalities, such as audio or facial cues, further restricted disambiguation in visually complex or information-dense scenarios.

\subsection{Findings from the User Survey}
A survey of 413 participants was conducted to contextualise the design requirements for automated name extraction within contemporary media consumption practices. Participants were recruited through social media platforms and online distribution channels, and the survey was administered as a structured online questionnaire comprising multiple-choice and Likert-scale items. The demographic profile was broad but skewed towards adults aged 18-34 and included a near-balanced distribution of male and female respondents, reflecting a general digital-news-consuming population rather than a domain-specific professional cohort.

The responses confirmed that graphical overlays remain an essential source of contextual information in news video content. A majority of participants, 59\%, reported difficulty in reading personal names presented on screen, with display speed, typographic variation, and low visual clarity identified as the primary causes. Many respondents noted that they frequently paused or rewound videos to verify an individual's identity, which reinforces the practical need for automated extraction systems that improve the accessibility of textual information in dynamic news environments.

Attitudes toward automation indicated cautious support. Approximately 62.2\% of respondents stated that their trust in an automated system would depend on the transparency of its operation and the credibility of the organisation responsible for its development. These findings highlight the importance of designing solutions that prioritise accountability, interpretability, and user confidence, particularly in journalistic contexts where factual accuracy and public trust are critical.

\section{Conclusion}
This study examined the automated extraction of personal names from news video graphics, a task that requires the coordinated operation of visual detection, text recognition, and linguistic interpretation. The proposed NGD was curated to reflect the stylistic variability of contemporary broadcast and social-media overlays. This supported the development of ANEP, a deterministic and modular framework that integrates object localisation, character extraction, and named-entity resolution within a transparent and auditable processing pipeline. A comparative evaluation against a generative multimodal baseline established a rigorous benchmark between deterministic auditability and the stochastic reasoning characteristic of generative multimodal systems.

The Gemini-based generative pipeline achieved the highest aggregate F1-score (84.18\%), while ANEP delivered balanced precision and recall (79.9\% and 74.4\%) and provided explicit data lineage and verifiable decision paths. The underlying detector exhibited strong cross-domain robustness (95.8\% mAP@0.5), indicating suitability for deployment across heterogeneous news environments. A complementary user survey also highlighted transparency as a decisive factor for public trust, which reinforces the need for interpretable systems in journalistic and accountability-focused applications.

Although newer multimodal systems continue to emerge, their probabilistic end-to-end architectures retain fundamental limitations in transparency, reproducibility, and interpretability. These constraints restrict their suitability for journalistic and evidentiary use, even when they offer incremental gains in extraction accuracy. The results here show that generative models can surpass deterministic pipelines on headline metrics; however, the ANEP framework remains essential in scenarios requiring verifiable data lineage.

Current limitations include reduced performance for non-Latin scripts, susceptibility of OCR components to stylised or animated typography, and reliance on external APIs for specific processing stages. Future work should expand coverage to multilingual and cross-script contexts, including right-to-left and regional alphabets, and should integrate richer semantic metadata such as affiliations and speaker roles. Multimodal transformer architectures that jointly model audio, text, and vision represent a promising direction for real-time identity disambiguation. Efficiency-oriented techniques such as pruning, quantisation, and edge deployment may reduce computational overhead in live broadcast scenarios. Additional exploration of self-supervised and confidence-aware learning strategies could enhance robustness and scalability across large video archives.

Overall, the findings show that a transparent OD, OCR, and NER pipeline, evaluated against state-of-the-art generative baselines, can provide accurate, interpretable, and operationally viable solutions for automated name extraction. This advances broader objectives in media analysis, accessibility, and information accountability.

\noindent\rule{\linewidth}{0.3pt}

\bibliographystyle{IEEEtran}
\bibliography{refs}

\noindent\rule{\linewidth}{0.3pt}

\end{document}